\documentclass[a4paper,conference]{IEEEtran}
\IEEEoverridecommandlockouts                              
\usepackage[misc,geometry]{ifsym}
\usepackage{graphicx}
\usepackage{subfig}
\usepackage{multirow}
\usepackage{amsmath}
\usepackage{tabularx}
\usepackage{makecell}
\usepackage{amsfonts}
\usepackage{enumitem}
\usepackage{upquote}
\usepackage{xcolor}
\usepackage{calligra}
\usepackage{array}
\usepackage{xcolor}
\usepackage{breqn}
\usepackage{balance}
\usepackage{hyperref}
\newcolumntype{P}[1]{>{\centering\arraybackslash}p{#1}}

\begin{document}
%
\title{Identity Documents Authentication based on Forgery Detection of Guilloche Pattern}

\author{\IEEEauthorblockN{Musab Al-Ghadi, Zuheng Ming, Petra Gomez-Kr\"{a}mer and Jean-Christophe Burie}
\IEEEauthorblockA{L3i, La Rochelle University \\ Avenue Michel Crépeau, 17042 La Rochelle, France.\\
Email: \{musab.alghadi,zuheng.ming,petra.gomez,jean-christophe.burie\}@univ-lr.fr}}


%

\maketitle
\begin{abstract}
In cases such as digital enrolment via mobile and online services, identity document verification is critical in order to efficiently detect forgery and therefore build user trust in the digital world. 
In this paper, an authentication model for identity documents based on forgery detection of guilloche patterns is proposed.   
The proposed approach is made up of two steps: feature extraction and similarity measure between a pair of feature vectors of identity documents. 
The feature extraction step involves learning the similarity between a pair of identity documents via a convolutional neural network (CNN) architecture and ends by extracting highly discriminative features between them. While, the similarity measure step is applied to decide if a given identity document is authentic or forged. 
In this work, these two steps are combined together to achieve two objectives: 
(i) extracted features should have good anti-collision (discriminative) capabilities to distinguish between a pair of identity documents belonging to different classes, 
(ii) checking out the conformity of the guilloche pattern of a given identity document and its similarity to the guilloche pattern of an authentic version of the same country.
Experiments are conducted in order to analyze and identify the most proper parameters to achieve higher authentication performance. The experimental results are performed on the MIDV-2020 dataset.
The results show the ability of the proposed approach to extract the relevant characteristics of the processed pair of identity documents in order to model the guilloche patterns, and thus distinguish them correctly. The implementation code and the forged dataset are provided here (\href{https://drive.google.com/drive/folders/1WCzQrIcfklgUdHQjKtpNyWzKsEujgdUD?usp=sharing}{https://drive.google.com/id-FDGP-1})

\end{abstract}

%
\IEEEpeerreviewmaketitle

\section{Introduction}
With the explosion of internet and in an ever-more digital world that is governed by new regulations, the providers of digital services like registration and distributed services (via mobile) are keen to offer more secure and more straightforward digital identification systems. 
In addition, COVID-19 pandemic has created and accelerated a number of key universal shifts that are casting the procedure in which the customers are accurately identified when interacting with digital services or signing up to new accounts.
Across the world, governments are passing legislation to protect their documents and their citizens from counterfeits and frauds. The crimes like unpaid invoices or money laundering typically involve fake or stolen identity documents, and accounts tied to false or fraud identities.
Thus, confirming the legitimacy of an identity document such as a passport, a national identity card or even a driving license has become a core part of online security and digital on-boarding, and constitutes a significant challenge for businesses. 
In response, regulations for distributed registration and know your customer (KYC) force businesses to have more rigorous processes and secure systems for verifying the identity documents of their customers.
To combat any forgery style on identity document, governments incorporated a number of sophisticated security features in all issued identity documents. These features make it theoretically difficult, if not impossible, to produce counterfeited or forged identity documents \cite{Jung2021}. 
The security features include for example the optical variable devices (OVDs). OVDs are composed of several patterns, logos and fonts that appear and disappear according to the light and the viewer’s position. All of these features are difficult to reproduce accurately which make them good anti-counterfeiting features. Holograms, guilloche, optically variable ink, anti-scan patterns, watermarks, micro types, encoded data, and invisible fluorescent fibers are examples of security features\footnote{https://regulaforensics.com/en/knowledge-hub/glossary-documents}.

Accordingly, an automatic and rapid deep identity documents verification system to detect all forgery possibilities can be designed by examining the identity documents’ integrated security features \cite{Berenguel2019}.
Indeed, the mentioned security features help to design advanced technologies for identity document verification at different levels including facial verification \cite{Ming2017,Wu2019,Chinapas2019}, conformity of visual features/patterns \cite{Chernov2015,Ghanmi2018,Castelblanco2020,Ghanmi2021}, content coherence \cite{Centeno2019}, and finally quality features \cite{alGhadi2022}.
One of the interesting visual features/patterns is the printed template on the background of an identity document. This pattern is called guilloche and is defined as a geometrical pattern of computer-generated fine lines that are interlaced to form a unique shape \cite{Stepien1998,Usilin2011}. 
This pattern is an attractive feature to confirm the authenticity of a given identity document \cite{Usilin2011}. This may be achieved directly by checking out the conformity of the guilloche pattern at the background of the identity document and its similarity to the guilloche pattern of an authentic version from the same country.

This paper proposes a new authentication approach based on guilloche verification to confirm whether a user’s identity document is authentic or forged.
The solution is therefore based on designing an intelligent and precise verification solution, which manages to read the entire identity document and to recognize the guilloche pattern to check its similarity to other patterns of authentic identity document from the same country. The objective of the similarity check is to validate its authenticity or its rejection. Concretely, the approach is made up of two steps: (i) extracting highly discriminative features between a pair of identity documents via a CNN model, (ii) measure the similarity between the extracted features.



\section{Literature Review}

This section explores some of the most relevant state-of-the-art approaches for the identity document verification. These approaches are presented in this section according to the studied feature or the technologies used in each one. 

Fundamentally, the visual features of identity documents are the features that have been most studied in the literature to build authentication approaches for identity documents \cite{Chernov2015,Ghanmi2018,Castelblanco2020}.
In \cite{Chernov2015} the authors proposed a verification approach for the passport documents based on detecting and verifying the periodic patterns (logos) that are printed on the passport documents. The presence or absence of the periodic patterns on a given passport document is studied through \emph{k} peaks of the Fast Fourier Transform (FFT) components. Hence, the average information of the 8 neighbours of each peak is calculated and compared with a predefined threshold to discriminate between a genuine and a forged passport document. 
In \cite{Ghanmi2018} an authentication approach for identity documents based on conformity of visual features and patterns was also proposed. 
Here, the approach is based on the generation of a visual descriptor called Grid Color Connected Components Descriptor (Grid-3CD) from a set of visual features relevant enough to the color connected components of the processed identity document. The similarity between the descriptors of a genuine identity document and a query identity document is measured to discriminate whether the query identity document is genuine or forged. 
The approach of \cite{Castelblanco2020} is composed of two main modules for identity documents acquisition and verification. The first module is a pre-processing module, which localizes the identity document, while the second module was a specific classifier to verify the identity document type and its legitimacy. The classifier module started by extracting local and global features like grayscale histograms, hue and saturation differences (HSD), structural similarity score (SS), and histogram of oriented gradients (HOG) from a given identity document. Then, these features are inserted into the support vector machine (SVM) and random forest (RF) classifiers to test if the document is genuine or forged. 
Another solution for identity document authentication was designed based on the CNNs as in \cite{Ghanmi2021,Centeno2019}. In \cite{Ghanmi2021}, two CNN models called siamese and triplet CNN are adapted to design an approach for identity document verification. The role of those models is to extract the feature vectors of a pair of identity documents and then the similarity between these vectors are measured to decide if a given identity document is authentic or not. 
\cite{Centeno2019} used Siamese, Triplet, and PeleeNet CNN models to design a verification approach for Spain identity documents. Those two models are not end-to-end learning models; as they are concerned with specific regions of the document to analyze and not whole the identity document.
Furthermore, computer vision algorithms were also used to propose an efficient approach for identity documents verification like in \cite{Yoosuf2020}. The approach involved two steps: the first step used the oriented fast and rotated brief (ORB) method to localize the security features like seal, signature and stamp on the processed identify document, while the second step used optical character recognition (OCR) and local binary pattern (LBP) to extract some significant features from the processed identity document. The authenticity of the processed document is evaluated by matching the LBP as sliding window operations. 
The facial and personnel information verification of identity documents are also important elements to detect any forgery on the identity documents \cite{Wu2019}. 
In \cite{Wu2019} two main steps were involved: the first is the facial verification, which based on inception-ResNet face embedding (IRFE) and the second is the verification of the personal information using a morphology transformed feature mapping (MTFM) and a deep cascading multitasking framework (MTCNN). 
Recently, a reference hashing approach for quality detection of identity documents was proposed in \cite{alGhadi2022}. As the high frequency components of the FFT are more relevant to represent the details information of a given identity document and are highly affected by the scanning operation; the proposed approach worked on them to build a highly discriminative hash code for a given identity document. Hence, this approach helps to identify the source of the identity document and to define whether it is genuine or generated after a print-scan operation.\\
According to all of the mentioned and described approaches, we can synthesize some open issues and challenges in the domain of identity documents authentication. (i) Most of the proposed approaches require the existence of the original identity document (as a reference) to detect if a given identity document is genuine or forged. (ii) Most of the proposed approaches are evaluated on private datasets, this makes comparison of these approaches difficult and even impossible to evaluate their performances.
(iii) No forged (fake) public dataset for identity documents is available to allow researchers to improve and more easily advance their works in this domain. Actually, the proposed approach endeavours to cover these issues and challenges.
 
\section{Proposed Model}

The proposed forgery detection approach for efficient identity document verification is made up of two steps: feature extraction and feature similarity measure.
The first step involves extracting the highly discriminative features of a pair of identity documents via a CNN model called Siamese Neural Network (SNN).
While, the second step calculates the similarity between the extracted features vectors to predict whether the input pair of identity documents belongs to the same class or not. 
The following subsections describe in details these steps.


\subsection{Feature Extraction and Loss Function}

\begin{figure}[bp] 
\centering
\includegraphics[width=0.5\textwidth]{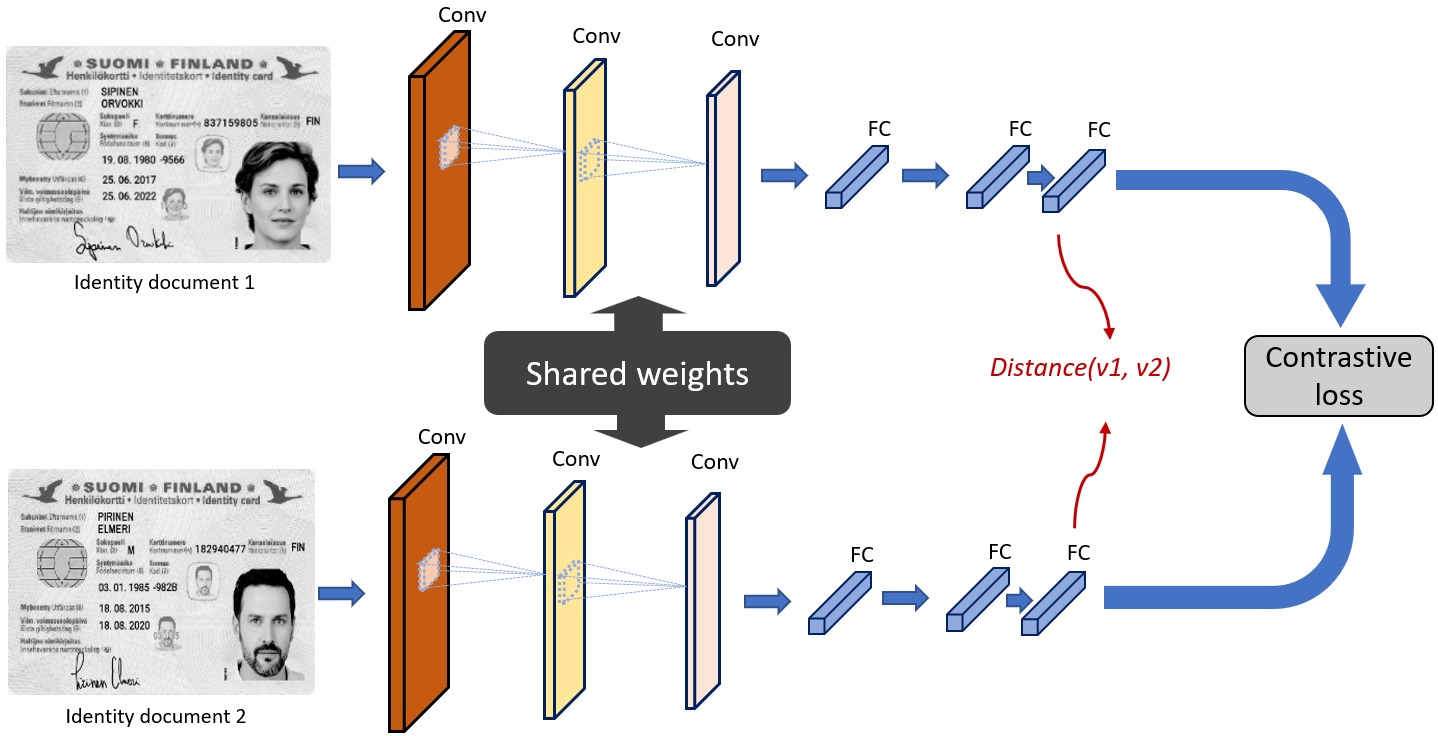}
\caption{Structure of the used SNN to extract the feature vectors.} 
\label{SNN}
\end{figure}

Actually, the SNN consists of a pair of neural networks, which are identical to each other and thus allows to process a pair of identity documents picked from the dataset to extract highly discriminative features between them \cite{Ghanmi2021,Dey2017}. The selection of SNN in this work comes due to three reasons \cite{Ghanmi2021,Dey2017}: (i) the fact that the SNN needs fewer training samples to classify/discriminate identity documents, (ii) it learns the semantic similarities rather than the differences like in other convolutional networks, and (iii) it mainly uses for verification of originality.
The general structure of the used SNN to extract the feature vectors is shown in Fig. \ref{SNN}, and its configuration is illustrated in Table \ref{SNN-conf}.

The left block in Fig. \ref{SNN} presents a pair of identity documents and the goal is to determine if these two identity documents belong to the same class or not. The middle block shows the convolutional network, which is the SNN in this work. The two subnetworks of SNN have the same convolutional (Conv) layers and parameters and mirror each other (i.e. if the weights in one subnetwork are updated, then the weights in the other is updated as well). The last layers in these subnetworks are typically the fully connected (FC) layers where we can compute the distance between the outputs and adjust the weights of the subnetworks accordingly. The right block then shows the loss function, which uses to optimize the network. 

The loss function calculates the distance between the outputs of the Conv layer to see if the SNN made the correct decision or not. The contrastive loss \cite{Chen2021,Hadsell2006} is the loss function used in this work and is given in equation \ref{eq:loss}: 

\begin{equation}
loss = c \times d^2 + (1 - c) \times max(m - d, 0)^2
\label{eq:loss}
\end{equation} 

\noindent where \emph{c} equals 1 if the pair of identity documents belongs to the same class, otherwise 0. The margin \emph{m} is arbitrary set to 2.0, while the distance \emph{d} expresses the similarity score between a pair of identity documents and is calculated according to equation~\ref{eq:Dis}:
 
\begin{equation}
\label{eq:Dis}
d = \sum^L_{l=1} \delta[v_{1l} \neq v_{2l}]
\end{equation}
\noindent where \emph{v$_1$} is the feature vector of the identity document 1 and \emph{v$_2$} is the feature vector of identity document 2 and each in length \emph{L}.

\begin{table}[tp]
\caption{Configuration of the SNN.}
\label{SNN-conf}
\centering
\tiny
\begin{tabularx}{\columnwidth}{p{1.2cm} P{1.3cm} P{1.2cm} P{0.3cm} P{0.3cm} P{0.65cm} P{1.2cm} }
\hline
Layer & Input shape & Feature maps & Filter & Stride & Padding & Activation function\\\hline
conv2d  & 1$\times$300$\times$300 & 4$\times$300$\times$300 &  3$\times$3 & 1$\times$1 & pad 1 & Hyperbolic tangent (tanh)\\
ReLU   & 4$\times$300$\times$300 & 4$\times$300$\times$300 &   &  &  & \\
BatchNorm2d   & 4$\times$300$\times$300 &  4$\times$300$\times$300&   &  &  & \\
conv2d  & 4$\times$300$\times$300 & 8$\times$300$\times$300 &  3$\times$3 & 1$\times$1 & pad 1 & Hyperbolic tangent (tanh)\\
ReLU   & 8$\times$300$\times$300 & 8$\times$300$\times$300 &   &  &  & \\
BatchNorm2d   & 8$\times$300$\times$300 &  8$\times$300$\times$300&   &  &  & \\
conv2d  & 4$\times$300$\times$300 & 8$\times$300$\times$300 &  3$\times$3 & 1$\times$1 & pad 1 & Hyperbolic tangent (tanh)\\
ReLU   & 8$\times$300$\times$300 & 8$\times$300$\times$300 &   &  &  & \\
BatchNorm2d   & 8$\times$300$\times$300 &  8$\times$300$\times$300&   &  &  & \\
FC  &  8$\times$300$\times$300 & 500 & & & &  \\
ReLU   & 500 & 500 &  &    &  & \\
FC  &  500 & 500 & &  &  &   \\
ReLU   & 500 & 500 &   &  &   & \\
FC  &  500 & 5 & & & &   \\ \hline
\hline
\end{tabularx}
\end{table}

It is worth noting that an SNN is a special type of CNNs where instead of building a learning model to classify inputs by learning their differences, it learns their similarities. Additionally, an SNN does not classify the inputs into certain categories or labels, rather it only finds out the distance between two inputs. If the inputs have the same class, then the network should learn the parameters (i.e. the weights and the biases) in such a way that they should produce a smaller distance between the two inputs, otherwise, the distance should be larger.

\begin{figure}[tp]
\centering
\includegraphics[width=0.48\textwidth]{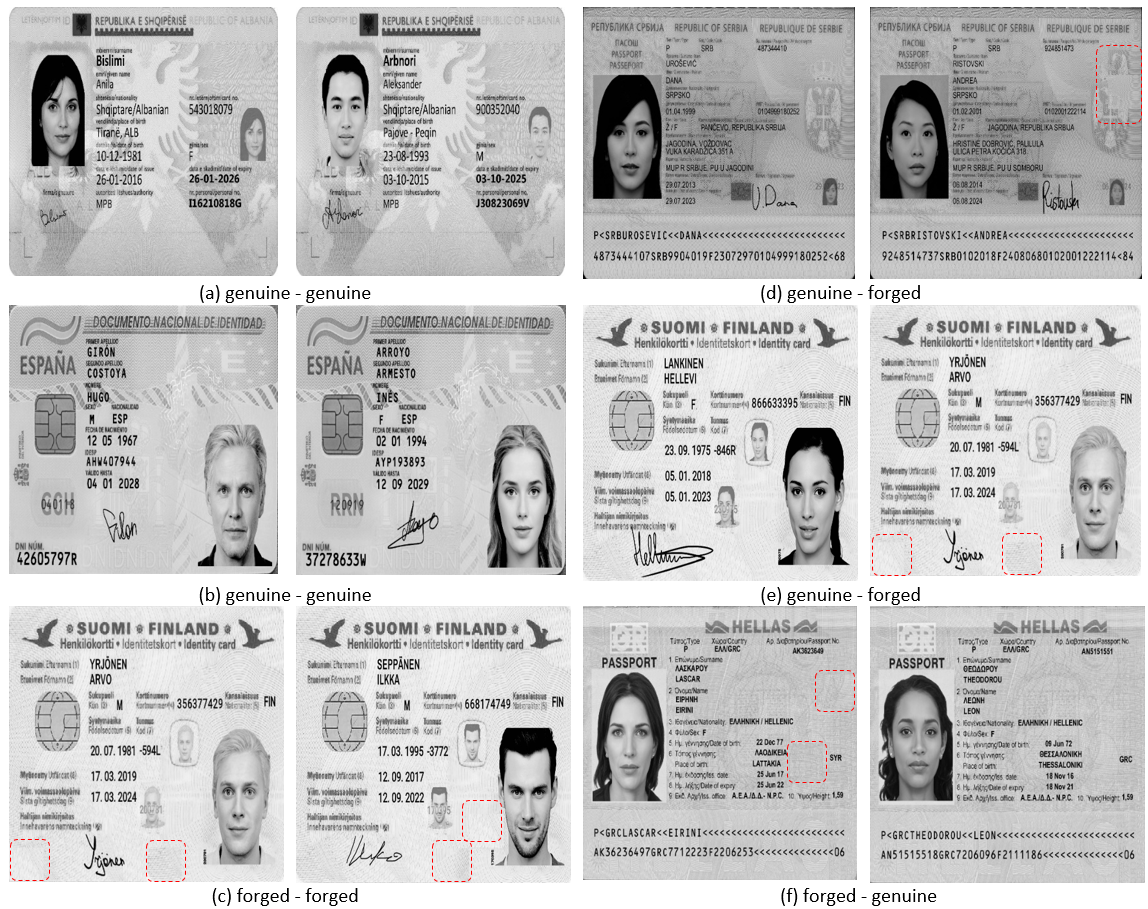}
\caption{Samples of similar pairs (a, b, c), and dissimilar pairs (d, e, f); red boxes present the forged locations.} 
\label{Samples}
\end{figure}

Since an SNN consists of two identical subnetworks, so we need to prepare pairs of identity documents as inputs. Hence, we have either similar pairs or dissimilar pairs of identity documents. A similar pair consists of two identity documents that belong to the same class (either to the genuine class or to the forged class) and of the same country, while a dissimilar pair consists of two identity documents that belong to different class (one identity document belongs to the genuine class and the other belongs to the forged class), and also of the same country.
Samples of similar pairs and dissimilar pairs of identity documents are showed in Fig. \ref{Samples}. 

At the end of this step, the proposed approach selects the obtained 5 values from the SNN and use them as discriminative features to train the network based on the contrastive loss value.
When the network has been trained, the obtained discriminative features are passed into the test step to measure the similarity between two identity documents and then decide if the two identity documents belong to the same class or not.

\subsection{Similarity Measure}

\begin{figure}[bp]
\centering
\includegraphics[width=0.48\textwidth]{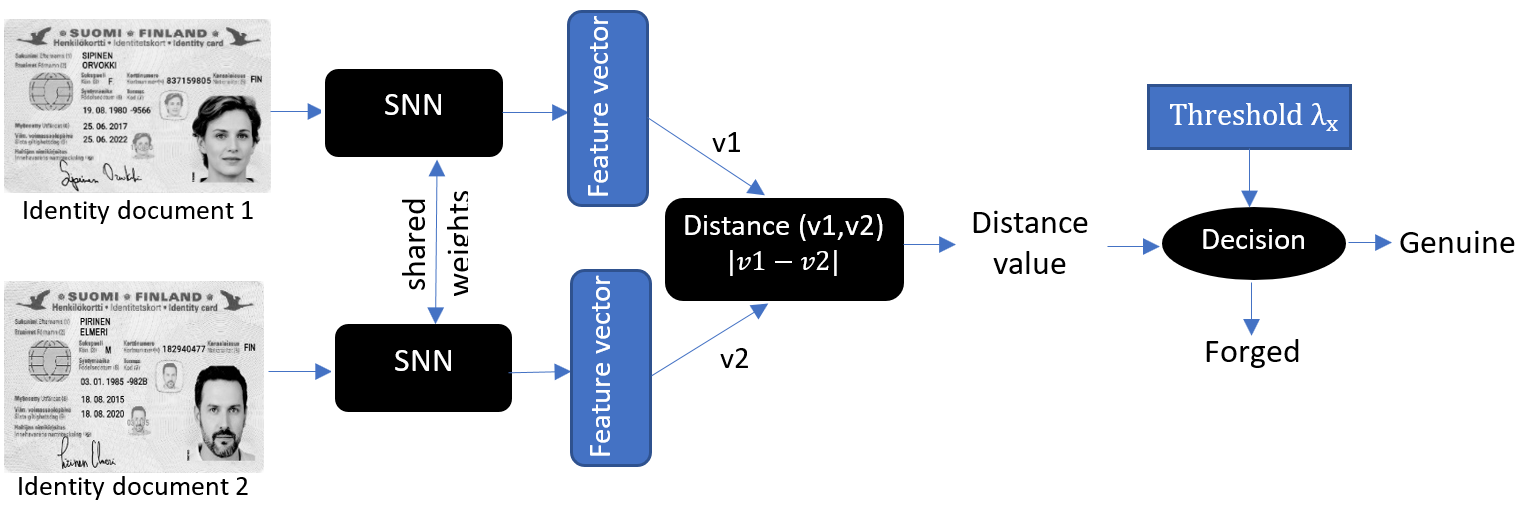}
\caption{Similarity measure and decision making scheme.} 
\label{sim}
\end{figure}

The proposed approach achieves two functions in this step: the first function is a similarity measure and the second is a decision making.
The first function calculates the distance between the two extracted feature vectors of the two identity documents and results in a similarity score as output.
Afterwards, the decision whether a query identity document is genuine or forged takes place by comparing the similarity score with a pre-defined threshold $\lambda$.
The structure of the similarity measure step is illustrated in Fig.~\ref{sim}.

To set up the threshold $\lambda$, the distances distribution between the set of similar and dissimilar pairs need to be studied (an example introduces in subsection \ref{subsection:parameter}). The distances between the set of similar and dissimilar pairs are calculated according to equation \ref{eq:Dis}.
Hence, if the obtained distance is less than a pre-defined threshold $\lambda$, then the two identity documents are considered as similar pair; otherwise, the two identity documents are considered as dissimilar pair.
In real world experiments, it is worth noting that we need to define one input identity document as genuine in order to predict whether the second identity document (i.e. the query) is genuine or forged.

\section{Experimental Results}

This section details the information about the dataset, the experiments setting, the evaluation metrics, and finally the experimental results of the proposed approach.

\subsection{Identity Documents Dataset and Generating a Forged Dataset}

To evaluate the performance of the proposed approach, the MIDV-2020 dataset \cite{Bulatov2021} is used.
The MIDV-2020 dataset consists of 1000 video clips, 2000 scanned images, and 1000 photos of 1000 unique dummy (template) identity documents with their annotations to read the ground-truths. These samples comprise identity documents and passports for 10 different countries.

Specifically, the 1000 unique dummy identity documents and the 1000 upper-right scanned identity documents are selected to represent the genuine dataset. While, the forged dataset is generated from the genuine dataset.
To this end, a copy-move tampering operation of specific zones in the genuine identity document of a specific country is applied. These zones are selected as follows:
For each country, one identity document is selected and divided into a set of non-overlapping blocks of size N$\times$N (as example 16$\times$16, 32$\times$32, 64$\times$64, etc.). Then, the zones (blocks) which visually contain only the guilloche pattern and do not have any foreground components like logos, photos, signatures, or textual information are selected to be candidate zones for applying copy-move operations to generate a forged dataset.

\begin{figure}[bp]
\centering
\includegraphics[width=0.48\textwidth]{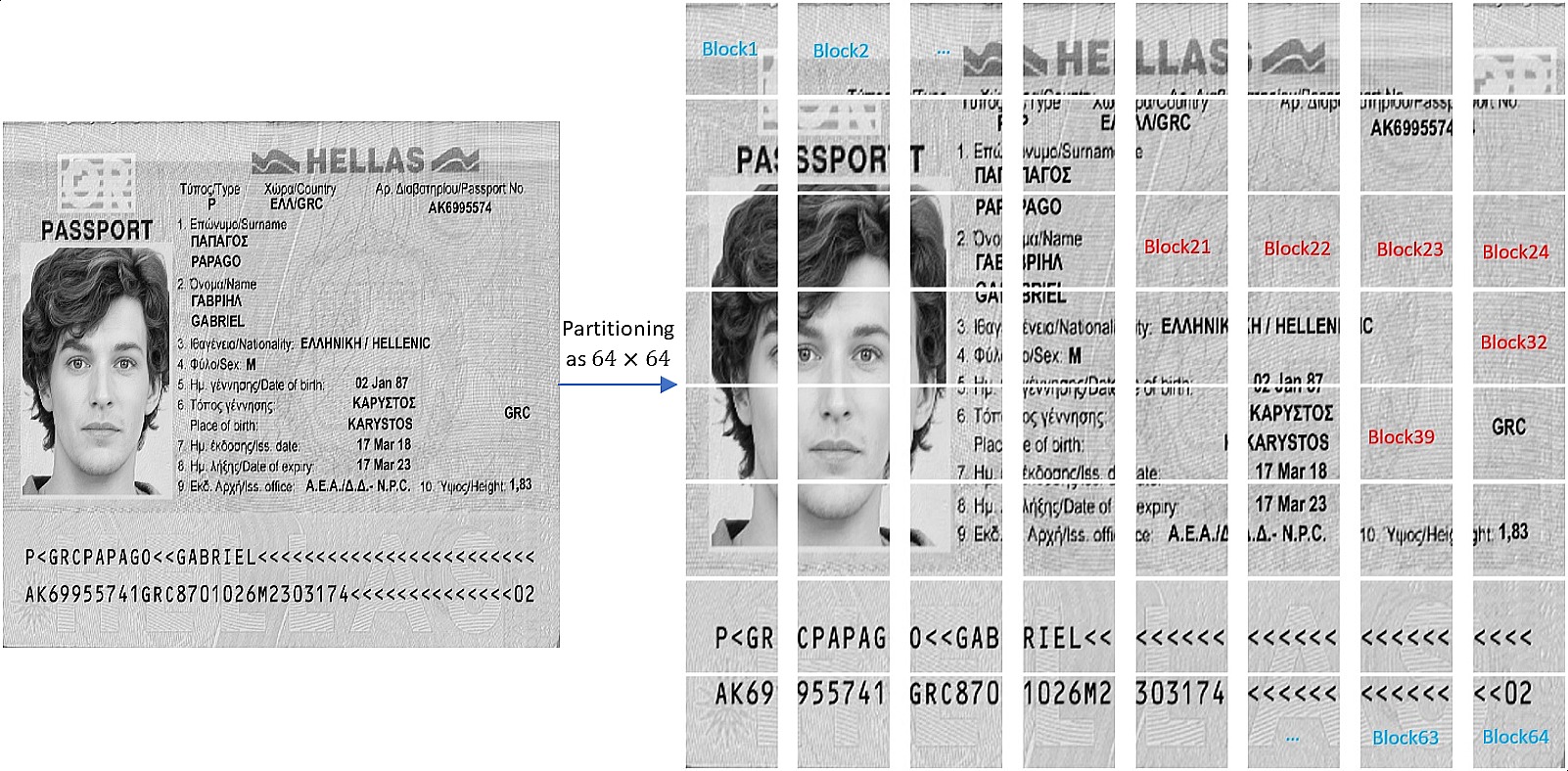}
\caption{Partitioning process and selecting candidates zones for copy-move operation.} \label{candidatesBlocks}
\end{figure}

Fig. \ref{candidatesBlocks} illustrates the partitioned process of one identity document of size 512$\times$512 into blocks of size 64$\times$64, and how we select the candidate zones for a copy-move operation. Indeed, the red annotated blocks \{block$_{21}$, block$_{22}$, block$_{23}$, block$_{24}$, block$_{32}$, block$_{39}$\} are selected as a set of zones for applying a copy-move operation to generate a forged dataset of the given country. Whereas all of these zones visually contain only information about the guilloche pattern. Fig. \ref{Samples} presents several examples of the generated forged identity documents, where the copy-move operation has been applied on the red surrounded zones.

The reason why we decided to apply this methodology to create the forged dataset is based on the fact that forged documents are often created with copy-move operations. Fraudster usually hide their modifications by reusing some patterns of the genuine document. For example, if the fraudulent name is shorter than the original one, the unwanted text is hidden by a copy-move operation. And area without any information are good candidates for this operation. However, the candidate area contains a guilloche pattern. The copy-move operation will create irregularities in the global pattern that our approach wants to detect.

As a consequence, the genuine dataset which consists of 2000 identity documents and the forged dataset which also consists of 2000 identity documents are both used to test the performance of the proposed approach.
It is worth noting that the 1000 unique dummy identity documents were created from Wikimedia (https://commons.wikimedia.org) Commons 
as template samples \cite{Bulatov2021}. While, the 1000 upper-right scanned identity documents were created by scanning the template samples using Canon LiDE 220 and Canon LiDE 300 scanners with a resolution of 2480$\times$3507 \cite{Bulatov2021}.
2/3 of the total identity document samples (2000 genuine and 2000 forged) are randomly selected to be used as training dataset and the rest is used as testing dataset.

\subsection{Experiments Setting}

The training implementation has been carried out on a local server (28 CPUs, 128 Go RAM, 4 $\times$ GPU Nvidia RTX 2080Ti), the batch size is 4, and the number of epochs is 500. The Adam optimizer is used, where the learning rate (lr) is equal to 5e-5 and the weight decay is equal to 0. While, our testing implementation has been running on HP laptop, Intel(R), Core(i7).

\subsection{Evaluation Metrics} 

The true acceptance rate (TAR), the false rejection rate (FRR), and the false acceptance rate (FAR) metrics are used to evaluate the discrimination performance of the proposed approach. Actually, TAR, FRR, and FAR are calculated according to the equations \ref{AR}, as reported in \cite{Ouyang2015}.

\begin{equation}
\begin{split}
\footnotesize
\label{AR}
TAR(\lambda) = \frac{x_1(d<\lambda)}{X_1},
FRR(\lambda) = \frac{x_2(d<\lambda)}{X_1},
\\
FAR(\lambda) = \frac{x_3(d<\lambda)}{X_2} 
\end{split}
\end{equation}

\noindent where \emph{x$_1$} is the number of similar pairs of identity documents that are classified as similar pairs, 
\emph{x$_2$} is the number of similar pairs of identity documents that are classified as dissimilar pairs,
\emph{x$_3$} is the number of dissimilar pairs of identity documents that are classified into similar pairs, \emph{X$_1$} and \emph{X$_2$} correspond to the total number of similar pairs and dissimilar pairs of identity documents, respectively. $\lambda$ is the discrimination threshold to consider a pair of identity documents as similar or dissimilar. \emph{d} as defined in equation (\ref{eq:Dis}) is the similarity score between the feature vectors of the two processed identity documents. 
The receiver operating characteristics (ROC) curve is also used to evaluate the discrimination performance of the proposed approach on different thresholds. This curve represents the results of the \emph{TAR} and the \emph{FAR} with different thresholds.

\subsection{Parameter Determination}
\label{subsection:parameter}

\begin{figure}[!tp]
\centering
	\label{subfig:1}
	\includegraphics[width=0.5\textwidth]{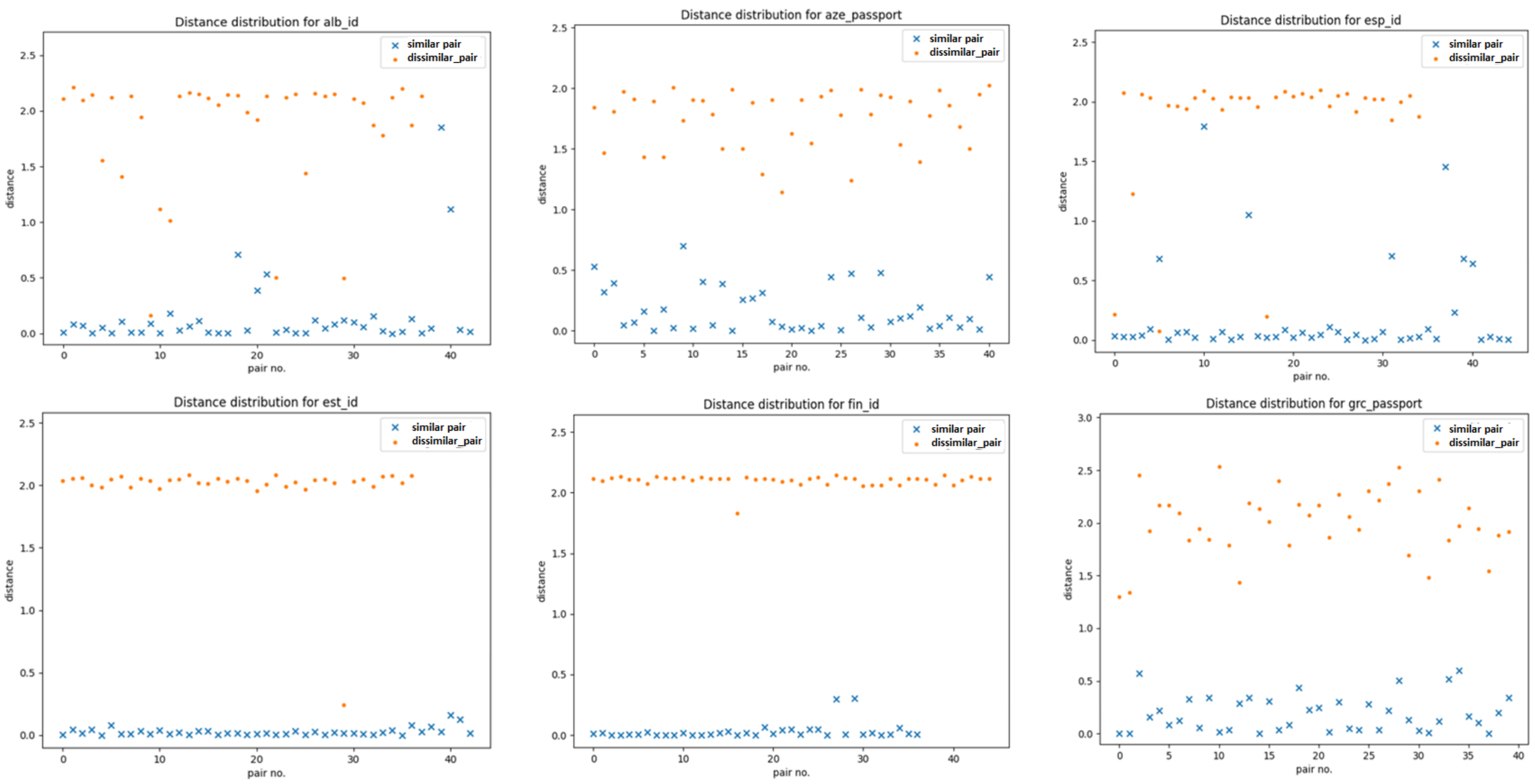}  

	\label{subfig:2}
	\includegraphics[width=0.5\textwidth]{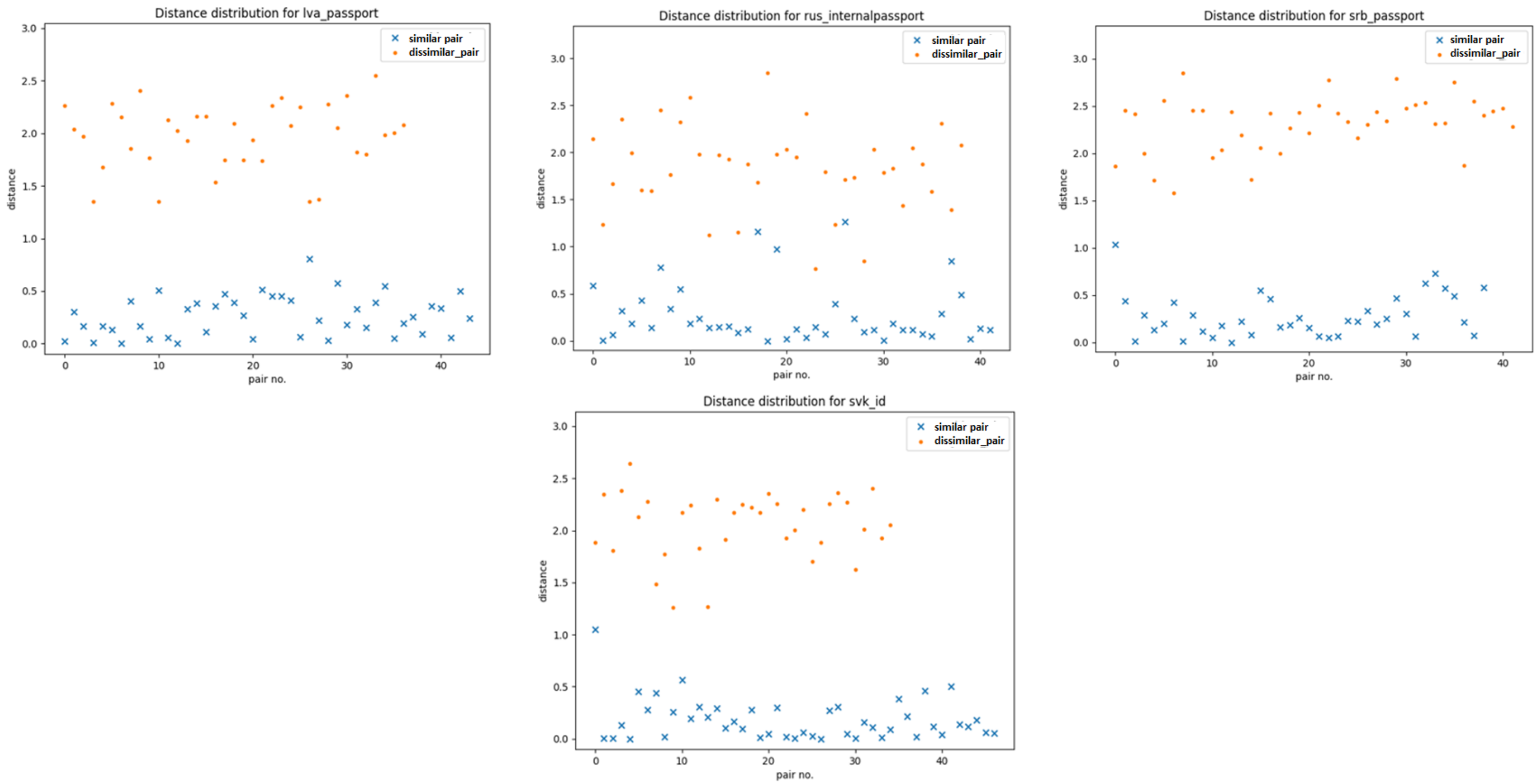}

\caption{Distances distributions between similar pairs (blue) and dissimilar pairs (orange) of the 10 countries of MIDV-2020.} 
\label{distributions}
\end{figure}

The discrimination threshold ($\lambda$) has a direct influence on the performance of the proposed approach and it needs to be determined.
It is worth noting that the discrimination threshold ($\lambda$) is calculated for each of the processed countries in MIDV-2020, that is to say 10 countries.
To do so, for each country the distances between the feature vectors of the processed pairs of identity documents are calculated.
Fig. \ref{distributions} presents as example of the distance distributions of 80 processed pairs of identity documents for the 10 countries.

From Fig. \ref{distributions} we can study the minimum and the maximum distances between the similar and dissimilar pairs of each country, and then define a discriminative threshold $\lambda$ for each country. Table \ref{parameter} shows the minimum and maximum distances for the similar and dissimilar pairs of each country, and accordingly we can define the range of possible thresholds for each country.
In case of overlap between the distances of similar and dissimilar pairs as example in \{Alb, Esp, Est, Rus\} countries, we select the narrow margin of distances which guarantee maximum discrimination between the similar and dissimilar pairs; otherwise, we select the wide margin of distances.
\begin{table}[bp]
\caption{Parameter determination.}
\label{parameter}
\centering
\scriptsize
\begin{tabularx}{\columnwidth}{p{1.5cm} |P{0.85cm} P{0.85cm} |P{0.85cm} P{0.85cm} |P{1.45cm} }
\hline
 & \multicolumn{2}{c|}{similar pairs} & \multicolumn{2}{c|}{dissimilar pairs} &  \\\hline
Country &min distance	&max distance	&min distance	&max distance	&Threshold   \\\hline
Alb (Albania) &0.0	&1.75	&0.20	&2.50	&[0.50 -- 1.00]   \\
Aze (Azerbaijan)&0.0	&0.73	&1.20	&2.45	&[0.75 -- 1.15]   \\
Esp	(Spain)&0.0	&1.82	&0.12	&2.11	&[0.75 -- 1.20]   \\
Est	(Estonia)&0.0	&0.23	&0.26	&2.12	&[0.25 -- 1.80]    \\
Fin	(Finland)&0.0	&0.42	&1.82	&2.18	&[0.48 -- 1.75]  \\
Grc	(Greece)&0.0	&0.65	&1.26	&2.53	&[0.65 -- 1.20]  \\
Iva	(Latvia)&0.0	&0.89	&1.37	&2.56	&[0.90 -- 1.30]   \\
Rus	(Russia)&0.0	&1.28	&0.89	&2.88	&[0.85 -- 1.10]  \\
Srb	(Serbia)&0.0	&1.02	&1.55	&2.92	&[1.05 -- 1.50]   \\
Svk	(Slovakia)&0.0	&1.03	&1.22	&2.61	&[1.05 -- 1.20]   \\
\hline \hline
\end{tabularx}
\end{table}
From table \ref{parameter} we can observe that the distances between the similar pairs and the dissimilar pairs of 6 countries \{Aze, Fin, Grc, Iva, Srb, Svk\} are completely discriminated. In cases of \{Alb, Esp, Est, Rus\} countries, there is a possibility to have small FARs because the distances between the similar pairs and the dissimilar pairs are not fully discriminated. However, we have a wide range of values to select one as a robust threshold to discriminate between the similar pairs and the dissimilar pairs of each of the mentioned countries. 
Consequently, each country will have one reference threshold $\lambda$ for forgery detection uses.

\subsection{Performance Test}

\begin{figure}[!bp]
\centering
\includegraphics[width=0.5\textwidth]{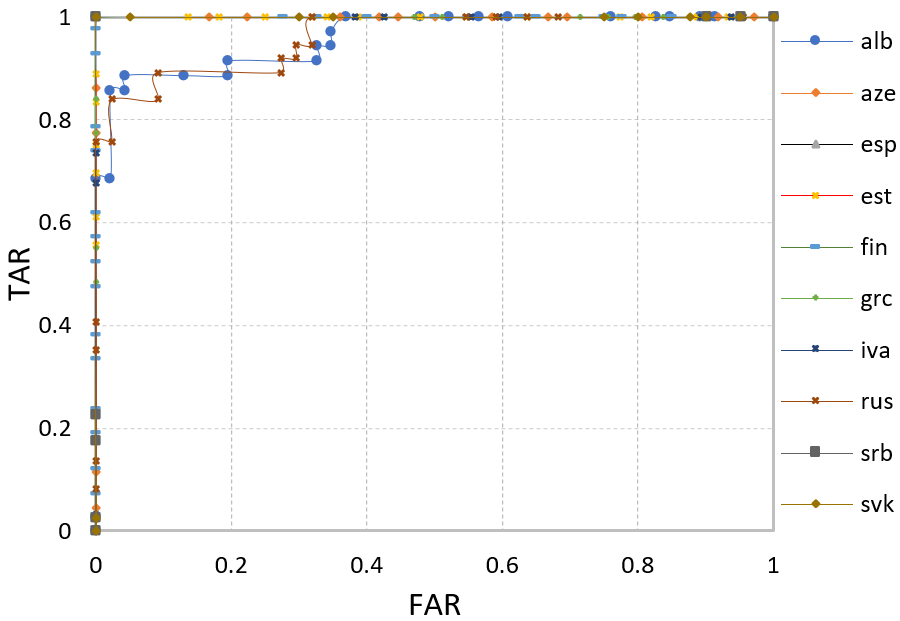}
\caption{ROC curves of the proposed approach of the 10 countries of MIDV-2020.} \label{ROC}
\end{figure}

Table \ref{performance} shows the performance test of the proposed approach in terms of TAR, FRR, and FAR of the 10 countries in MIDV-2020 under the specific threshold $\lambda$. 
Table \ref{performance} shows a very interesting results, where the TAR(s) exceeds 92\%, while both FRR(s) and FAR(s) are less than 8\% in all countries.
More accurately, the TAR(s) for 6 countries \{Aze, Est, Fin, Grc, Iva, Svk\} are equal to 1.0 and the FAR(s) are almost equal to 0.0.
Moreover, Fig. \ref{ROC} shows the TAR and FAR results with varying discriminative thresholds $\lambda$ for all countries of MIDV-2020. The presented ROC curves demonstrate the robustness and the discriminative capability of the proposed approach to distinguish pairs of identity documents correctly.

\begin{table}[!bp]
\caption{Performance test of the proposed approach in terms of TAR, FRR, and FAR under a given threshold $\lambda$.}
\label{performance}
\centering
\scriptsize
\begin{tabularx}{\columnwidth}{p{1.5cm} P{1.3cm} P{1.3cm} P{1.3cm} P{1.3cm} }
\hline
Country &Threshold $\lambda$	&TAR    &FRR	&FAR \\\hline
Alb     &1.0	    &0.95   &0.05	&0.08	\\
Aze 	&1.0	    &1.0    &0.0	&0.04	\\
Esp	    &1.0	    &0.93   &0.07	&0.08	\\
Est	    &0.5	    &1.0    &0.0	&0.02	\\
Fin	    &0.95	    &1.0    &0.0	&0.0	\\
Grc	    &0.95	    &1.0    &0.0	&0.0	\\
Iva	    &1.0	    &1.0    &0.0	&0.0    \\
Rus	    &1.0	    &0.92   &0.08	&0.05	\\
Srb	    &1.1	    &0.98   &0.02	&0.0	\\
Svk	    &1.1	    &1.0    &0.0	&0.0	\\
\hline \hline
\end{tabularx}
\end{table}

\subsection{Comparative Study and Discussion}

To the best of our knowledge our work is the first attempt to design a forgery detection model for identity documents w.r.t. guilloche pattern and no experiments or even results have been reported on the MIDV-2020 dataset in the literature in the domain of identity document authentication. Therefore, a comparative study is not possible from a practical point of view. Indeed, this paper introduces the baselines for any future work in this domain.

However, the novelty of our work can be demonstrated by making a systematic comparison between our work and the other works in \cite{Ghanmi2021,Centeno2019}, which are the mostly relevant approaches to our proposal, mainly in the following aspects:
(i) Generality of learning scheme: both \cite{Ghanmi2021} and \cite{Centeno2019} are not end-to-end learning models, as those models are concerned with specific regions of the document to analyze. While, our work is an end-to-end learning model regardless which region of the document to analyze. Hence, the proposed approach authenticates a given identity document at a whole, not merely zones targeted in the identity document like in \cite{Ghanmi2021} and \cite{Centeno2019};
(ii) Requirement of the original document (as a reference) to accomplish the verification process:
both \cite{Ghanmi2021} and \cite{Centeno2019} require the original documents to accomplish the verification process. While, our work requires one genuine
document that belongs to the same country and it is selected randomly;
(iii) Simplicity based on the number of used CNNs: the work of \cite{Ghanmi2021} used 1 CNN model (Siamese or Triplet) and the work of \cite{Centeno2019} used 3 CNN models (Siamese, Triplet, and PeleeNet). Similar to the work in \cite{Ghanmi2021}, our work used 1 CNN model (Siamese).
(iv) Threshold used: the work of \cite{Ghanmi2021} used a fixed threshold for each region (each targeted region has a specific threshold), while \cite{Centeno2019} used the Softmax, but is targeted to a specific region. More effectively, our work used a dynamic and global threshold for any type of identity documents that belongs to the same country.
(v) Types of processed documents:
the work of \cite{Ghanmi2021} checks the performance of their model on four types of French identity documents, and \cite{Centeno2019} test their model on banknotes of 12 countries  and only on the identity documents of Spain. More generally, our work checks the authentication of the passports and the identity cards of 10 countries.
(vi) Creation of new dataset: the work of \cite{Ghanmi2021} used a private dataset, no forged documents are introduced publicly, and \cite{Centeno2019} adds 11 new country banknotes. While, our work introduces publicly 2000 forged identity documents of 10 countries of MIDV-2020.

\section{Conclusion}
An authentication approach for identity documents based on the forgery detection of guilloche patterns has been proposed in this paper. The proposed approach comprised two steps, the first step used the SNN as a robust enough convolutional network to extract the highly discriminative features between an authentic identity document and a forged identity document of the same country. Whilst, the second step performed the similarity calculation between the extracted features, in order to confirm whether a given identity document is authentic or forged.
Experiments were conducted on the MIDV-2020 dataset and on a generated forged dataset, and the obtained results proved the capability of the proposed approach to extract the relevant characteristics of the processed pair of identity documents in order to model the guilloche pattern, and distinguished them correctly.

\section*{ACKNOWLEDGMENT}

This work is financed in the framework of FUI IDECYS+ (project n° : DOS0098984/00).

%

\bibliographystyle{spiebib}
\balance
\bibliography{bibliography}

\begin{thebibliography}{10}

\bibitem{Jung2021}
C.~Jung, G.~Kim, M.~Jeong, and et~al., ``Metasurface-driven optically variable
  devices,'' {\em Chemical Reviews}~{\bf 121}(21), pp.~13013--13050, 2021.

\bibitem{Berenguel2019}
A.~Berenguel, O.~R. Terrades, J.~Llados, and C.~Canero, ``E-counterfeit: A
  mobile-server platform for document counterfeit detection,'' in {\em 2019
  International Conference on Document Analysis and Recognition (ICDAR)},
  pp.~15--20, 2019.

\bibitem{Ming2017}
Z.~Ming, J.~Chazalon, M.~M. Luqman, M.~Visani, and J.~C. Burie, ``Simple
  triplet loss based on intra/inter-class metric learning for face
  verification,'' in {\em {IEEE} International Conference on Computer Vision
  Workshops (ICCVW)},  pp.~656--1664, 2017.

\bibitem{Wu2019}
X.~Wu, J.~Xu, J.~Wang, Y.~Li, W.~Li, and Y.~Guo, ``Identity authentication on
  mobile devices using face verification and {ID} image recognition,'' {\em
  Procedia Computer Scienc}~{\bf 162}, pp.~932--–939, 2019.

\bibitem{Chinapas2019}
A.~Chinapas, Polpinit, N.~Intiruk, and K.~R. Saikaew, ``Personal verification
  system using {ID} card and face photo,'' {\em International Journal of
  Machine Learning and Computing}~{\bf 9}, pp.~407--412, 2019.

\bibitem{Chernov2015}
T.~Chernov, V.~M. Kliatskine, and D.~Nikolaev, ``A method of periodic pattern
  detection on document images,'' in {\em Digital Library of the European
  Council for Modelling and Simulation},  pp.~1--5, 2015.

\bibitem{Ghanmi2018}
N.~Ghanmi and A.~Awal, ``A new descriptor for pattern matching: Application to
  identity document verification,'' in {\em 13$^{th}$ IAPR International
  Workshop on Document Analysis Systems (DAS)},  pp.~375--380, IEEE, 2018.

\bibitem{Castelblanco2020}
A.~Castelblanco, J.~Solano, C.~Lopez, E.~Rivera, L.~Tengana, and M.~Ochoa,
  ``Machine learning techniques for identity document verification in
  uncontrolled environments: A case study,'' in {\em Mexican Conference on
  Pattern Recognition},  pp.~271--281, Springer, 2020.

\bibitem{Ghanmi2021}
N.~Ghanmi, C.~Nabli, and A.~Awal, ``Checksim: A reference-based identity
  document verification by image similarity measure,'' in {\em Barney Smith
  E.H., Pal U. (eds) Document Analysis and Recognition--ICDAR 2021 Workshops},
  pp.~422--436, Springer, 2021.

\bibitem{Centeno2019}
A.~B. Centeno, O.~R. Terrades, J.~L. Canet, and C.~C. Morales, ``Recurrent
  comparator with attention models to detect counterfeit documents,'' in {\em
  2019 International Conference on Document Analysis and Recognition (ICDAR)},
  pp.~1332--1337, 2019.

\bibitem{alGhadi2022}
M.~Al-Ghadi, P.~Gomez-Kr\"{a}mer, and J.~C. Burie, ``Checkscan: A reference
  hashing for identity document quality detection,'' in {\em Fourteenth
  International Conference on Machine Vision (ICMV 2021)},   {\bf 12084},
  pp.~141--148, International Society for Optics and Photonics, SPIE, 2022.

\bibitem{Stepien1998}
P.~Stepien, R.~Gajda, and A.~Marszalek, ``Guilloche in diffractive optically
  variable image devices,'' in {\em Optical Security and Counterfeit Deterrence
  Techniques II},  pp.~231--236, SPIE, 1998.

\bibitem{Usilin2011}
S.~Usilin, D.~Nikolaev, and D.~Sholomov, ``Guilloche elements recognition
  applied to passport page processing,'' in {\em 8$^{th}$ Open German - Russian
  Workshop Pattern Recognition and Image Understanding},  pp.~1--5, 2011.

\bibitem{Yoosuf2020}
M.~S. Yoosuf and R.~Anitha, ``Forgery document detection in information
  management system using cognitive techniques,'' {\em Journal of Intelligent
  and Fuzzy Systems} , pp.~8057--8068, 2020.

\bibitem{Dey2017}
S.~Dey, A.~Dutta, J.~I. Toledo, S.~K. Ghosh, J.~Llados, and U.~Pal, ``Signet:
  Convolutional siamese network for writer independent offline signature
  verification,'' {\em Pattern Recognition Letters}~{\bf 1}, pp.~1--7, 2017.

\bibitem{Chen2021}
C.~Chen, Y.~Xie, S.~Lin, R.~Qiao, J.~Zhou, X.~Tan, Y.~Zhang, and L.~Ma,
  ``Novelty detection via contrastive learning with negative data
  augmentation,'' in {\em Thirtieth International Joint Conference on
  Artificial Intelligence (IJCAI)},  pp.~606--614, 2021.

\bibitem{Hadsell2006}
R.~Hadsell, S.~Chopra, and Y.~LeCun, ``Dimensionality reduction by learning an
  invariant mapping,'' in {\em {IEEE} Computer Society Conference on Computer
  Vision and Pattern Recognition (CVPR'06)},   {\bf 2}, pp.~1735--1742, 2006.

\bibitem{Bulatov2021}
K.~Bulatov, E.~Emelianova, D.~Tropin, and et~al., ``{MIDV}-2020: A
  comprehensive benchmark dataset for identity document analysis,'' {\em
  Computer Vision and Pattern Recognition, arXiv:2107.00396} , pp.~1--17, 2021.

\bibitem{Ouyang2015}
J.~Ouyang, G.~Coatrieux, and H.~Shu, ``Robust hashing for image authentication
  using quaternion discrete fourier transform and log-polar transform,'' {\em
  Digital Signal Processing}~{\bf 41}, pp.~98--109, 2015.

\end{thebibliography}

\end{document}